\documentclass{article}

\usepackage{arxiv}
\usepackage{multirow}
\usepackage[labelformat=parens]{subcaption}
\usepackage{colortbl}
\usepackage{xcolor}
\usepackage[utf8]{inputenc} 
\usepackage[T1]{fontenc}    
\usepackage{hyperref}       
\usepackage{url}            
\usepackage{booktabs}       
\usepackage{amsfonts}       
\usepackage{nicefrac}       
\usepackage{microtype}      
\usepackage{lipsum}
\usepackage{graphicx}
\usepackage{enumerate}
\usepackage[ruled, vlined]{algorithm2e}
\usepackage{float}

\title{Degradation Self-Supervised Learning for Lithium-ion Battery Health Diagnostics}

\author{
  Jie-Chung Chen \\
  Department of Materials Science and Engineering\\
  National Yang Ming Chiao Tung University\\
  Taiwan \\
  \texttt{jackson29.en11@nycu.edu.tw} \\
}

\begin{document}
\maketitle
\begin{abstract}
Health evaluation for lithium-ion batteries (LIBs) typically relies on constant charging/discharging protocols, often neglecting scenarios involving dynamic current profiles prevalent in electric vehicles. Conventional health indicators for LIBs also depend on the uniformity of measured data, restricting their adaptability to non-uniform conditions. In this study, a novel training strategy for estimating LIB health based on the paradigm of self-supervised learning is proposed. A multiresolution analysis technique, empirical wavelet transform, is utilized to decompose non-stationary voltage signals in the frequency domain. This allows the removal of ineffective components for the health evaluation model. The transformer neural network serves as the model backbone, and a loss function is designed to describe the capacity degradation behavior with the assumption that the degradation in LIBs across most operating conditions is inevitable and irreversible. The results show that the model can learn the aging characteristics by analyzing sequences of voltage and current profiles obtained at various time intervals from the same LIB cell. The proposed method is successfully applied to the Stanford University LIB aging dataset, derived from electric vehicle real driving profiles. Notably, this approach achieves an average correlation coefficient of 0.9 between the evaluated health index and the degradation of actual capacity, demonstrating its efficacy in capturing LIB health degradation. This research highlights the feasibility of training deep neural networks using unlabeled LIB data, offering cost-efficient means and unleashing the potential of the measured information.
\end{abstract}

\keywords{lithium-ion battery \and self-supervised learning \and electric vehicle}

\section{Introduction}

As one of the most versatile energy storage devices, lithium-ion batteries (LIBs) have become essential for applications such as electric vehicles (EVs)~\cite{sanguesa2021review} and mobile devices~\cite{liang2019review}. Despite their numerous advantages, LIBs undergo various side reactions \cite{che2023health, birkl2017degradation}, leading to irreversible capacity degradation and, in some cases, significant safety risks \cite{chen2021review}. Several characteristics have been identified as key indicators of battery degradation. For instance, an increase in internal resistance is considered a critical factor due to its association with the thickening of the solid electrolyte interphase (SEI) film or lithium dendrite formation~\cite{an2016state}. Besides, battery surface temperature rises due to excessive heat generation, which can result from electrolyte breakdown~\cite{rahman2024exploring} or overcharging/overdischarging~\cite{du2024side}. While extensive research has been devoted to understanding these phenomena, the complex degradation mechanisms of LIBs remain difficult to fully characterize and predict using empirical models or existing electrochemical theories alone.

With the rapid advancement of computing power, machine learning has emerged as a powerful tool for addressing complex and intractable problems. In the context of battery degradation, key metrics such as capacity, state of health (SOH), remaining useful life (RUL) and end of life (EOL) are commonly analyzed using supervised machine learning methods. Traditional supervised learning approaches often depend on human-selected features to achieve high accuracy~\cite{severson2019data}. However, recent advances in end-to-end deep learning, particularly in computer vision~\cite{chai2021deep} and sequential data processing~\cite{lim2021time}, have enabled accurate battery health estimation~\cite{fan2020novel} and life prediction~\cite{hsu2022deep, lu2022battery} directly from raw sensor and operational data, eliminating the need for explicit feature engineering. A notable limitation of these existing studies is their dependence on sufficient labeled data obtained from cycle aging and reference performance tests. For example, several publicly available LIB datasets~\cite{dos2021lithium} provide comprehensive battery degradation metrics along with detailed in-cycle measurements of capacity, temperature, and internal resistance. However, acquiring such high-quality data is often impractical for large-scale, real-world applications due to the substantial time and financial costs involved~\cite{lombardo2021artificial}.

Despite the abundance of operating data from lithium-ion batteries, much of it lacks the labeled annotations required for supervised learning. To address this limitation, self-supervised learning (SSL)~\cite{rani2023self}, which has been successfully applied in computer vision tasks and large language models, offers a potential solution. SSL leverages pretext tasks to train deep neural networks without relying on explicit semantic annotations, thereby reducing the need for manual data labeling. Based on their objectives, SSL methods can be broadly categorized into generative and contrastive approaches. Generative SSL typically employs an encoder-decoder architecture to reconstruct artificially masked input features. By recovering missing information from battery-related features, the model learns high-quality representations that enhance the generalization performance of downstream tasks such as state of charge (SOC)\cite{hannan2021deep} and SOH estimation\cite{wang2024lithium}. On the other hand, contrastive SSL focuses on distinguishing between similar and dissimilar data points. This method involves applying transformations or sampling techniques to generate multiple variants of a source feature, bringing similar variants closer together in an embedding space while pushing dissimilar ones aparts~\cite{luo2024remaining}. Both approaches aim to find the most effective task to pretrain model, minimizing the cost of downstream task fine-tuning. However, previous studies on battery degradation using SSL remain limited in practical usability, as the pre-trained model is non-functional without fine-tuning for specific downstream tasks. Moreover, most of these approaches assume static battery operating conditions, restricting their applicability in dynamic environments such as EVs and regulation reserves.

In this paper, inspired by the irreversibility of battery life, we propose a novel data-driven health estimation approach called degradation SSL. Unlike existing methods that rely on well-structured features, degradation SSL is highly adaptable and can be applied to diverse scenarios using only the chronological order of input features. Furthermore, fine-tuning for downstream tasks is not a necessary step, as the pretext task is inherently linked to battery health status. As a result, the model's output provides meaningful preliminary health estimations without additional supervised learning. To the best of our knowledge, degradation SSL is the first data-driven battery health estimation method capable of making inferences entirely independent of labeled annotations, offering a flexible and highly generalizable approach for a wide range of battery diagnostic applications.

\section{Methods}
\label{sec:method}

\subsection{Data description and preprocessing}

\begin{figure}[H]
  \centering
  \includegraphics[height=!,width=1\linewidth,keepaspectratio=true]{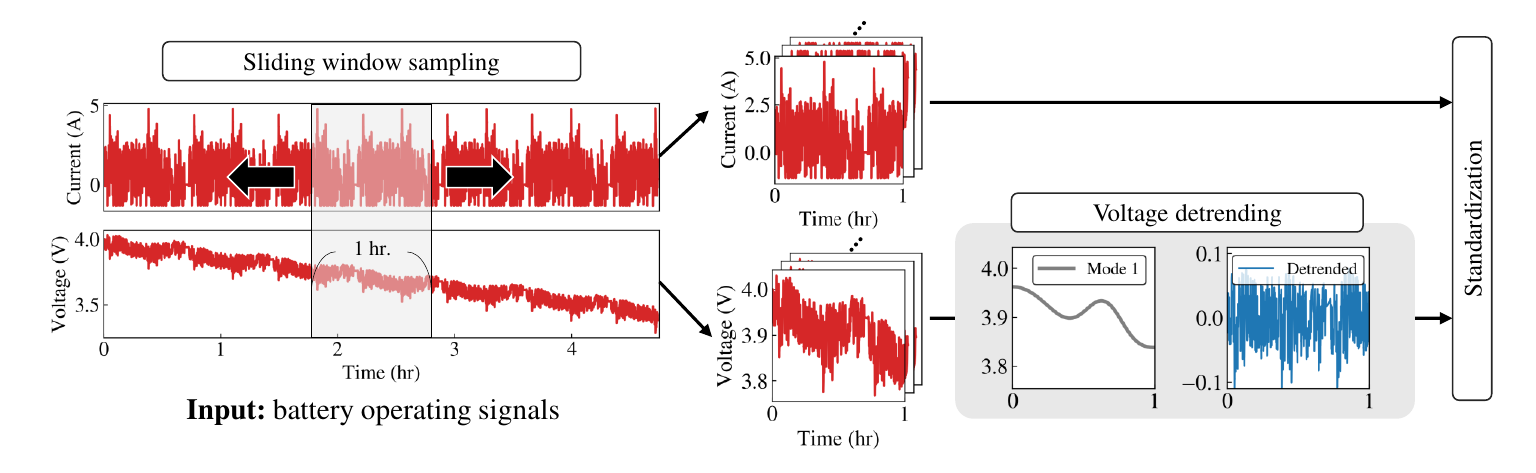}
  \caption{Preprocessing workflow of proposed method.}
  \label{fig:preprocess}
\end{figure}

The Stanford dataset \cite{pozzato2022lithium} contains aging data for 10 INR21700-M50T NMC battery cells under dynamic operating condition over a span of 23 months. During the aging cycles, the battery cells were charged following a CC-CV protocol and discharged based on an EV driving pattern derived from the Urban Dynamometer Driving Schedule (UDDS), which is commonly used for simulating city driving conditions. Diagnostic tests, including capacity test, Electrochemical Impedance Spectroscopy (EIS), and Hybrid Pulse Power Characterization (HPPC), were conducted periodically. Due to impedance abnormalities, two of the cells are excluded from this paper. Thus, the remaining 8 battery cells were utilized for the training and evaluation of the proposed framework.

In this study, only the voltage and current signals recorded during the UDDS discharge were employed as input features. As depicted in Figure \ref{fig:preprocess}, to ensure uniformity and manage input dimensions, a sliding window of 1-hour length randomly selects a patch from the UDDS discharge process as the representative sample for each cycle. During the training procedure, the input patches are resampled every epoch, thereby augmenting the diversity of training data. 

Distribution shift is a significant challenge in time series data analysis. Models designed for time-series tasks often suffer from the discrepancy among input sequences. For instance, the statistical properties of voltage signals during battery operation can be greatly influenced by the state of charge, leading to unstable data distributions. To address this problem, we adopted the empirical wavelet transform (EWT) \cite{gilles2013empirical} technique implemented in the research by Al-Greer et al. \cite{al2023capacity}, as a potential solution. EWT is an multiresolution analysis method, enabling adaptive decomposition for a target signal based on the characteristics in frequency domain. The implementation of EWT involves the following steps:
\begin{enumerate}[(i)]
    \item Derivation of the Fourier spectrum $\hat{f}(\omega)$ from the raw time series signal $f(t)$. 
    \item Assuming that $\hat{f}(\omega)$ can be segmented into at least $N$ modes, the boundaries of these modes can be determine as follows: identify the first $N$ local maximum in $\hat{f}(\omega)$ and locate $N-1$ midpoints between two adjacent maximum as boundaries. By including the starting and ending points of the spectrum, $N+1$ boundaries can be established.
    \item The detected $N+1$ boundaries are employed to build a total of $N$ filters through Littlewood-Paley and Meyer wavelets \cite{daubechies1992ten}.
    \item $N$ different modes are decomposed by applying corresponding filter to $\hat{f}(\omega)$. 
\end{enumerate}  
Instead of directly standardizing the input signals, we use EWT to pre-process the voltage data by detrending. Generally, the mode corresponding to the lowest frequency band captures the overall signal trend. Hence, we eliminate this mode from the raw signal to isolate the fluctuation of the voltage curve.

\subsection{Self-supervised degradation learning framework}
\begin{figure}[H]
  \centering
  \includegraphics[height=!,width=1\linewidth,keepaspectratio=true]{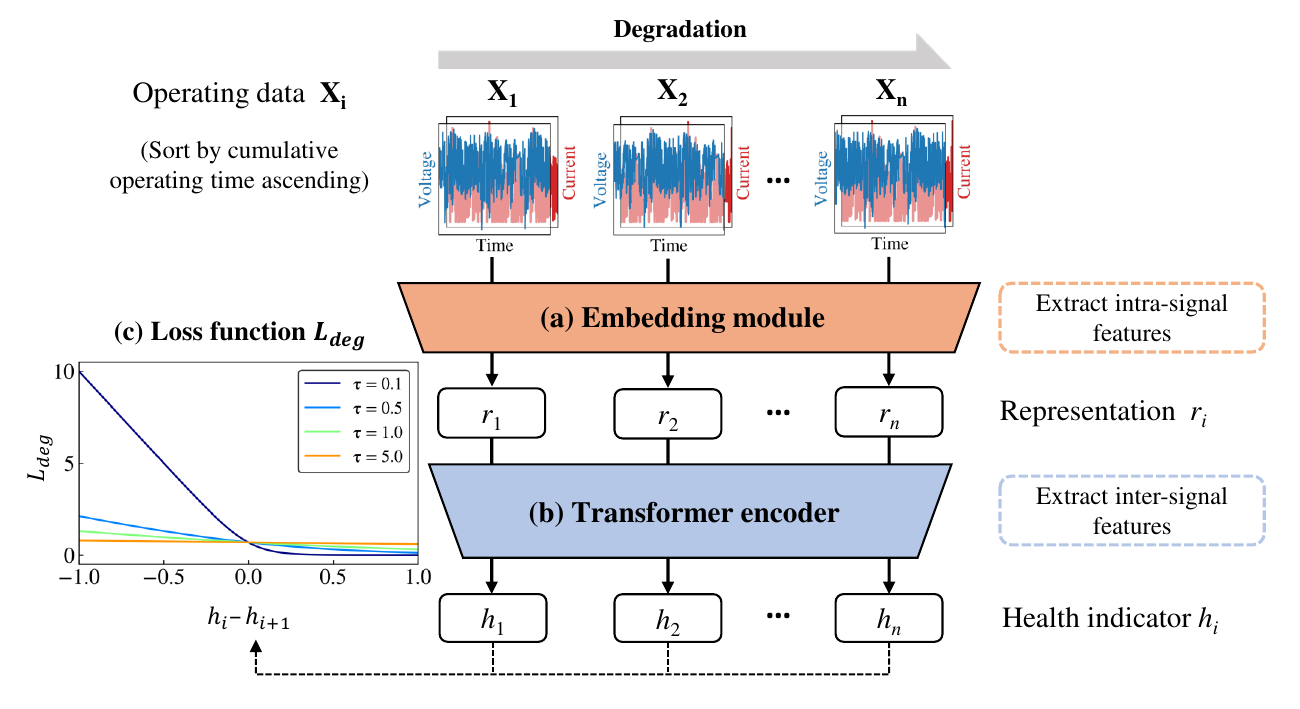}
  \caption{Overview of the proposed self-supervised degradation learning framework}
  \label{fig:ssl}
\end{figure}
The degradation of lithium-ion batteries is typically inevitable. Even though some special treatments enabling battery rejuvenation have been reported in the literature \cite{liu2021dynamic}, such instances are still uncommon in practical applications. Thus, in this work, we assume that the battery health status is progressively deteriorated. This allows the proposed neural network to estimate the battery health status based on unlabeled operational time-series data, such as discharge current and voltage, through a pretext task. The architecture of the neural network is shown in Figure \ref{fig:ssl} and detailed as follows.

\paragraph{Embedding module.}
As shown in Figure \ref{fig:ssl}(a), the embedding module is crucial for dimensionality reduction and feature extraction by mapping an input matrix of operating signals $\mathbf{X}_i$ to a representation $r_i$. Where $\mathbf{X}_i$ consists of time series data of both discharge voltage and current during a given time segment $i$; $r_i$ represents a one-to-one representation with $X_i$, containing the intra-signal features related to degradation. This module consists of convolutional layers with skip connections \cite{hek2016deep} and inception architectures \cite{szegedy2016rethinking} to effectively capture local temporal features in the time series data. Following the convolutional block, a fully-connected layer is adopted to aggregate these local features, generating the corresponding representations. Although the resulting representation $r_i$ has intra-cycle information of time segment $i$, it may lack sufficient information between other cycles. Thus, a further module is needed for robust health status assessment.

\paragraph{Transformer encoder.}
As shown in \ref{fig:ssl}(b), a transformer encoder is adopted as a decision-maker to assess health indicator $h_i$ for sequence data $\mathbf{X}_i$.  The core of the transformer model is self-attention mechanism, proficient in capturing complex and long-range dependencies inherent in sequential data. Specifically, it computes a weighted sum of values $V$, where the weights are determined by the compatibility between query $Q$ and key $K$ vectors. The compatibility scores are calculated through a dot product operation followed by softmax function, ensuring that the weights effectively reflect the importance of each sequence element with respect to the query. This process can be formulated as:

\begin{equation}
    {\rm Attention}(Q,K,V)={\rm softmax}(\frac{QK^T}{\sqrt{d_k}})V
\end{equation}

where $d_k$ denotes the dimensionality of $K$, serving as a normalization factor to maintain consistent dot product scales across different input sequences. Note that, in this study, $Q$, $K$, and $V$ vectors are derived from a trainable linear projection of a representation vector $r_i$.

\paragraph{Loss function.}
The assumption adopted in the proposed framework is that the health status of batteries decreases with operating time. Therefore, a differentiable metric which can serve as a loss function for self-supervised degradation learning is required to quantify the extent of health status degradation, such that:

\begin{equation}
    L_{deg}=-\sum_{i}\left({\rm log}\frac{D_i}{D_i+1}\right)
\end{equation}

where

\begin{equation}
    D_i={\rm exp}\left(\frac{h_i-h_{i+1}}{\tau}\right)
\end{equation}

where coefficient $\tau$ is inversely proportional to the sensitivity of $L_{deg}$ and is set as 0.1 in this study. As shown in Figure \ref{fig:ssl}(c), health status $h_i$ is expected to be greater than the subsequent $h_{i+1}$, i.e. $h_i-h_{i+1}>0$. This condition minimizes $L_{deg}$ and reduces its gradient. Conversely, when $h_i-h_{i+1}<0$, a significant loss is imposed to rectify the model. Thus, any inconsistency in health status degradation activates the penalty mechanism and increases the value of $L_{deg}$.

\subsection{Training procedure}
Mini-batch gradient descent is a strategy for efficiently updating model parameters by dividing the training data into multiple subsets. However, this approach requires input samples with consistent dimensionality, which presents a challenge when handling battery operating data with varying sequence lengths. To address this issue, we adopt a strategy where 20 cycles are randomly selected from each cell at the start of every iteration, ensuring consistency in sequence length within each mini-batch. Furthermore, this method enhances the diversity of input data, reducing the risk of model overfitting. For optimization, we used AdamW \cite{loshchilov2017decoupled} algorithm, which modifies the regularization term to improve generalization ability. The model is trained with a constant learning rate of 0.0001 for 1000 epochs.


\section{Results}

\subsection{Health status estimation}

\begin{figure}[h]
  \centering
  \includegraphics[height=!,width=1\linewidth,keepaspectratio=true]{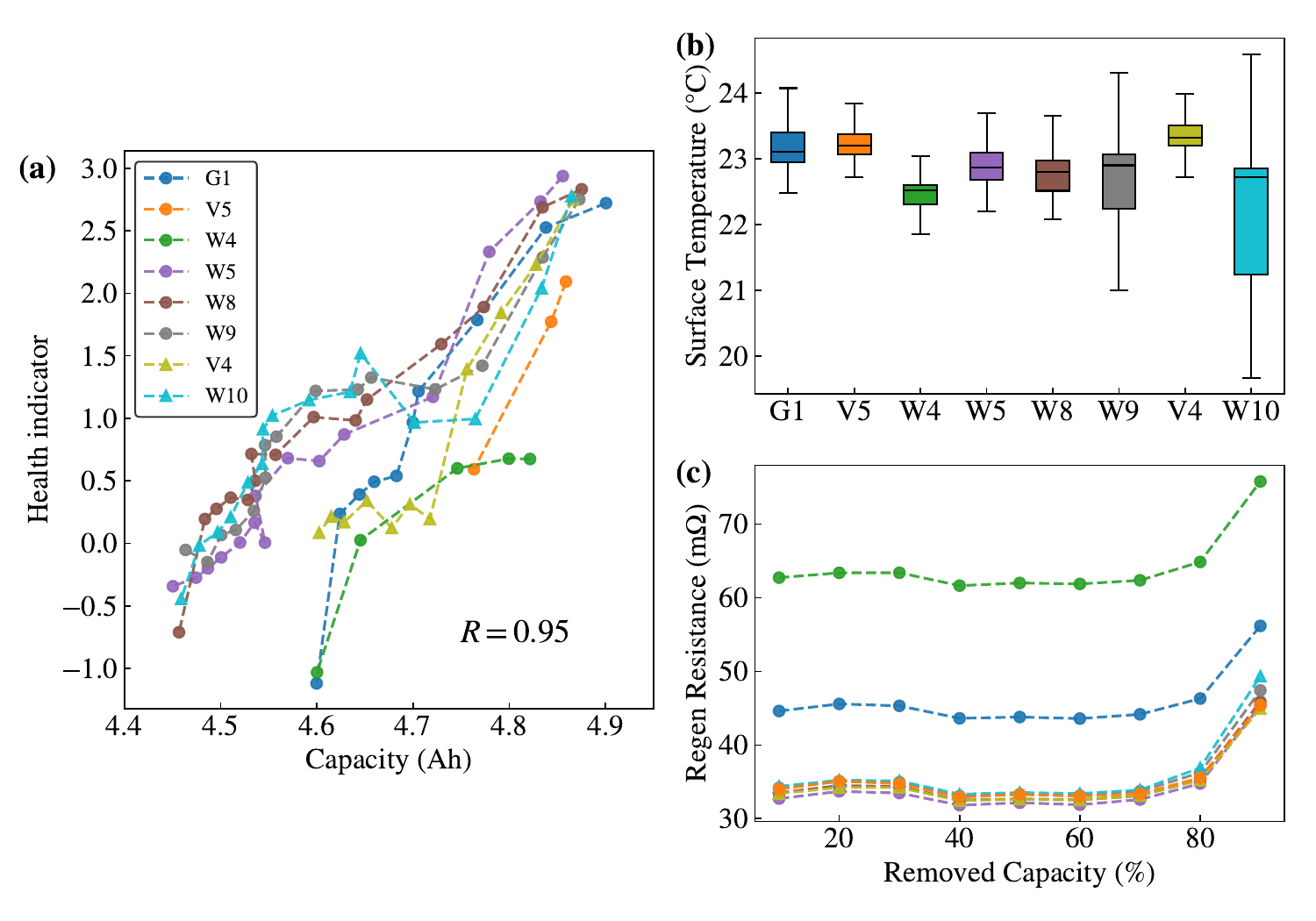}
  \caption{(a) Discharge capacity and the corresponding health indicator $h$ estimated by the proposed model for the 8 cells. (b) The distribution of surface temperature during the diagnostics tests. (c) The regen resistance calculated according to the HPPC at third diagnostic test.}
  \label{fig:single_cell}
\end{figure}

To show that the proposed self-supervised learning framework can successfully capture the degradation of batteries, one of the most highly correlated features with battery health status, i.e., discharge capacity, is selected for comparison with the corresponding health indicator $h$ estimated by the proposed model, as shown in Figure \ref{fig:single_cell}(a). However, it is worth to emphasize that the proposed model is trained without specific labeled data, and thus, the estimated $h$ may not solely reflect degradation characteristics in discharge capacity.

Eight selected cells (labeled as G1, V4, V5, W4, W5, W8, W9, and W10) are included, where V4 and W10 belong to the testing dataset, marked by a triangular symbol, while the remaining cells belong to the training dataset, marked by a circular symbol. The value of health indicator $h$ in Figure \ref{fig:single_cell}(a) is determined by the average of 10 randomly selected 1-hour segments of the voltage and current curves at the cycles where the measurement of discharge capacity was performed. The estimated health indicators exhibit a high correlation ($R=0.95$) with the discharge capacity. That is, a greater value of the health indicator typically indicates a higher discharge capacity in all cases. This can be clearly demonstrated by the cases of W5, W8, W9, and W10, where the $h$ of these cells (ranging from around 3 to -1) exhibits a nearly linear pattern consistent with discharge capacity (ranging from around 8.5 to 4.4 Ah). On the other hand, although the remaining four cells (G1, V5, W4, and V4) may not exhibit a perfect linear pattern and do not show specific ranges of the $h$ and discharge capacity values, the $h$ still decreases along with the discharge capacity in general. The reason why the trend of the estimated $h$ of G1, V5, W4, and V4 is not as ideal as that of W5, W8, W9, and W10 could be due to abnormal situations that occurred during the measurements. As illustrated in Figure \ref{fig:single_cell}(b), according to the experimental logs from the diagnostics tests \cite{pozzato2022lithium}, only the surface temperatures of cells G1, V4, and V5 are over 23°C. Moreover, the hybrid pulse power characterization tests at early cycle show that the calculated regen resistances of W4 and G1 are significantly higher than those of the others. It is noteworthy that these abnormal situations are expected to cause accelerated aging in the cells, and the proposed model can reveal this by assigning a relatively lower $h$ value, even when their discharge capacity values were still greater than 4.6 Ah.

\begin{figure}[h]
  \centering
  \includegraphics[height=!,width=1\linewidth,keepaspectratio=true]{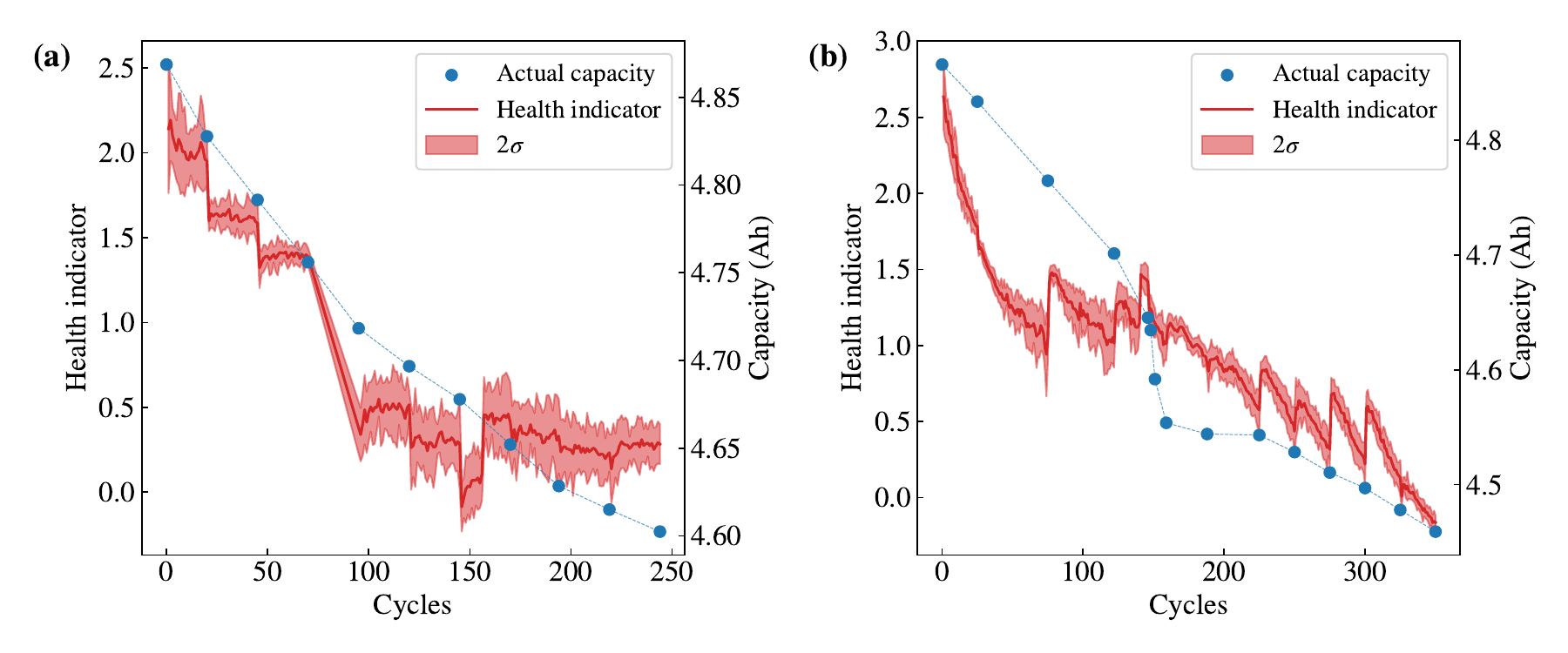}
  \caption{The averaged estimated $h$ of each cycle and the corresponding capacity of (a) V4 and (b) W10 cells.}
  \label{fig:testing_cells}
\end{figure}

Now consider practical electric vehicle operating scenarios. The battery management system typically only acquires data from the current cycle of the battery cells, while the complete historical data of the cells is absent. Thus, we propose utilizing the existing operating records of other cells as reference samples to estimate the health indicator based on limited data of the target cell. Specifically, we leverage 10 cycles from each cell in the training set as references to estimate the health indicator of the cells in the testing set (V4 and W10). Then, we randomly extract a 1-hour segment from those 10 cycles of each reference cell and another 1-hour segment from the target testing cell. The 10 segments from the reference cells can provide the operating information that is absent in the target testing cell. These 11 segments are then fed into the proposed model to estimate $h$ of the testing cell at that cycle. Figure \ref{fig:testing_cells} presents the averaged $h$ results of each cycle of V4 and W10 cells based on the above-mentioned approach. The red shaded area, labeled as 2$\sigma$, represents the distribution range of plus and minus two standard deviations, which is derived from the various $h$ values obtained through multiple random segment samplings within a cycle. It is remarkable that even when only 1 cycle from the testing cells is used at a time, i.e., no information from other cycles in the same cell is considered, the trend of the health indicator decreases over time as expected, showing the great applicability and practicability of the proposed model. Also, it is of interest to note that there are several abrupt jumps in the health indicator, which typically occurred at the cycles where discharge capacity measurements were performed (blur points in the Figure \ref{fig:testing_cells}). This phenomenon may be attributed to the calendar aging process between the cycle aging experiments~\cite{reichert2013influence}. Based on the date and time logs in the dataset \cite{pozzato2022lithium}, the average rest period between cycle aging tests for these cells is approximately 60 days, which is sufficient to induce calendar aging. Similar to the abnormal situations mentioned in the previous paragraph, the effects of calendar aging, which cannot be revealed by the discharge capacity, can be captured by the proposed model.

\subsection{Representation visualization}
\begin{figure}[H]
  \centering
  \includegraphics[height=!,width=1\linewidth,keepaspectratio=true]{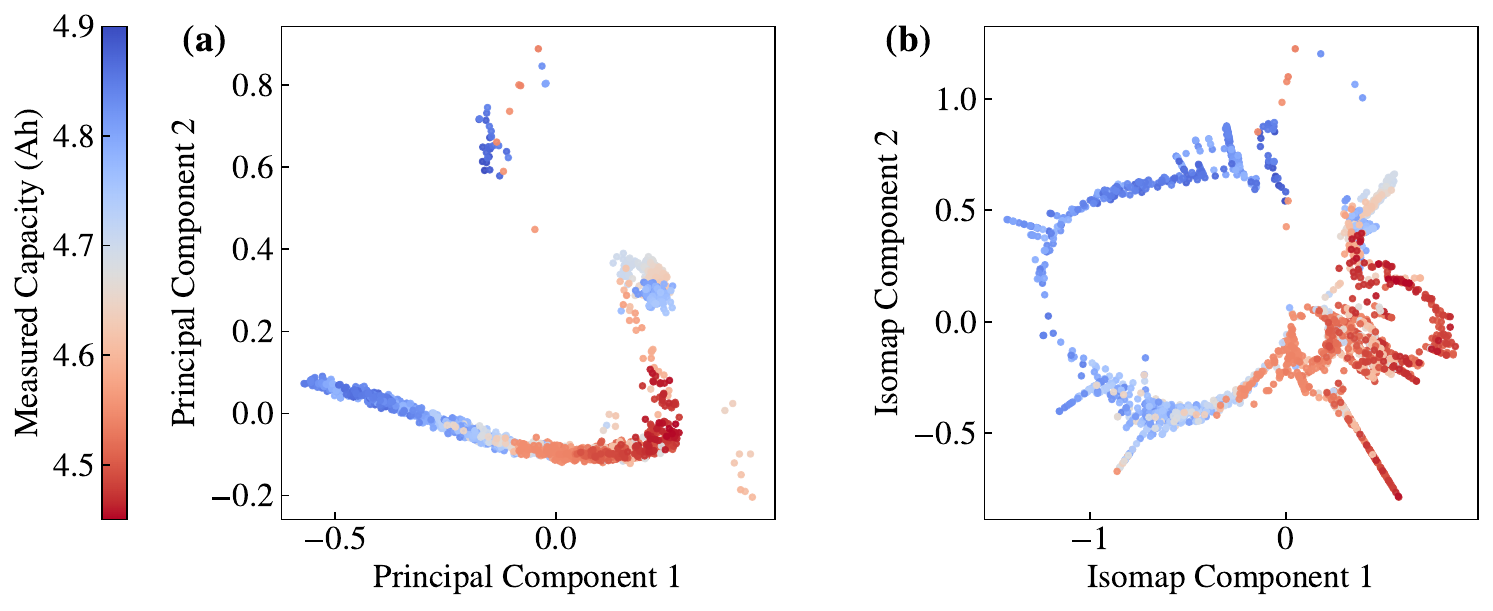}
  \caption{Visualization of the representations extracted by embedding module using (a) PCA and (b) Isomap algorithms}
  \label{fig:repres}
\end{figure}

We further visualize the representation manifold in 2D latent space using dimensionality reduction algorithms, namely principal component analysis (PCA) and isometric mapping (Isomap), to investigate the extracted representations from the operating signal, as illustrated in Figure \ref{fig:repres}(a) and (b), respectively. In the PCA algorithm, the covariance matrix of the dataset is computed, and the projection of the representative features that maximizes data variation in a linear relationship is determined. The Isomap algorithm calculates the geodesic distance between data points, retaining the distances between features after dimensionality reduction in the high-dimensional space. Note that each data point in Figure \ref{fig:repres} was extracted from a randomly selected 1-hour segment within the cycle during which capacity tests were performed, with the color gradient, ranging from red to blue, indicating the transition from low to high capacity measured in each cell. Both visualizations reveal clear distinctions among signal segments with varying capacities, highlighting a noticeable transition from high to low capacity along the curve in the latent space. The color gradient further emphasizes the relationship between the signal’s latent representations and capacity. 

\subsection{Ablation experiment}
\begin{table}[H]
    \begin{subtable}[t]{.5\linewidth}
        \centering
        \begin{tabular}{m{15mm}<{\centering} m{15mm}<{\centering} m{15mm}<{\centering}}
            \toprule
            case & $R$ & $L_{deg}$ \\
            \midrule
            \rowcolor{black!10}
            detrend & \textbf{0.93} & \textbf{0.50}\\
            none & 0.84 & 0.62 \\
            \bottomrule
        \end{tabular}
        \caption{\textbf{Preprocessing}}
    \end{subtable}
    \begin{subtable}[t]{.5\linewidth}
      \centering
        \begin{tabular}{m{15mm}<{\centering} m{15mm}<{\centering} m{15mm}<{\centering}}
            \toprule
            case & $R$ & $L_{deg}$ \\
            \midrule
            \rowcolor{black!10}
            CNN &  \textbf{0.93} &  \textbf{0.50} \\
            FC & 0.20 & 0.69 \\
            \bottomrule
        \end{tabular}
        \caption{\textbf{Embedding module}}
    \end{subtable}
    \begin{subtable}[t]{.5\linewidth}
      \centering
      \vspace{0.1cm}
        \begin{tabular}{m{15mm}<{\centering} m{15mm}<{\centering} m{15mm}<{\centering}}
            \toprule
            dim & $R$ & $L_{deg}$ \\
            \midrule
            128 & \textbf{0.93} & 0.51 \\
            \rowcolor{black!10}
            256 & \textbf{0.93} & \textbf{0.50} \\
            512 & 0.91 & 0.52 \\
            \bottomrule
        \end{tabular}
        \caption{\textbf{Model width}}
    \end{subtable} 
    \hfill
    \begin{subtable}[t]{.5\linewidth}
      \centering
      \vspace{0.1cm}
        \begin{tabular}{m{15mm}<{\centering} m{15mm}<{\centering} m{15mm}<{\centering}}
            \toprule
            depth & $R$ & $L_{deg}$ \\
            \midrule
            2 & \textbf{0.94} & \textbf{0.50} \\
            \rowcolor{black!10}
            4 & 0.93 & \textbf{0.50} \\
            8 & 0.93 & \textbf{0.50} \\
            \bottomrule
        \end{tabular}   
    \caption{\textbf{Model depth}}
    \end{subtable} 
    \caption{Ablation experiments on the Stanford dataset. Default settings are marked in gray.}
    \label{table:ablation}
\end{table}
To evaluate the effectiveness of key components in the proposed self-supervised framework, we conducted ablation experiments by comparing the correlation coefficient ($R$) and the loss function ($L_{deg}$). First, we assessed the role of the detrending process in data preprocessing, as shown in Table \ref{table:ablation}(a). The results indicate that without detrending, $R$ decreases by 0.09, while $L_{deg}$ increases by 0.12, demonstrating its crucial impact on model performance. This effect can be attributed to the fact that voltage curves at high and low SoC levels exhibit significant variations due to battery aging. However, in practical applications, obtaining voltage curves within a specific SoC range is often infeasible, as maintaining a battery at extreme SoC levels for data collection is impractical. Instead, we hypothesize that voltage fluctuations caused by rapid current changes contain essential information for health estimation. By removing the voltage-SoC trend and amplifying these fluctuations, we enhance the model’s ability to capture meaningful degradation patterns, ultimately improving its predictive performance.

Table \ref{table:ablation}(b) examines the impact of different embedding module architectures, specifically comparing convolutional neural networks (CNNs) with fully connected (FC) layers. The results reveal that models relying solely on FC layers exhibit significantly poor performance. This suggests that CNNs effectively capture spatial patterns in voltage fluctuations, which may be too complex for self-attention mechanisms to process without prior transformation. Given that voltage fluctuations contain critical information for health estimation, an effective embedding module is essential for extracting meaningful representations. These findings indicate that the design of the embedding module is likely more influential than the transformer encoder itself, as proper feature extraction facilitates better downstream learning.

In Table \ref{table:ablation}(c) and (d), we analyze the impact of transformer encoder width and depth. Our findings indicate that a shallow architecture is sufficient for learning degradation. This can be attributed to the nature of degradation self-supervised learning (SSL), which resembles a sorting problem and is inherently less complex than tasks in natural language processing or computer vision. As capturing high-level semantic information is not required, heavy self-attention mechanisms are unnecessary. Thus, a lightweight model architecture is adequate for addressing the degradation problem.

\subsection{Knowledge transfer to downstream task}

\begin{figure}[H]
  \centering
  \includegraphics[height=!,width=1\linewidth,keepaspectratio=true]{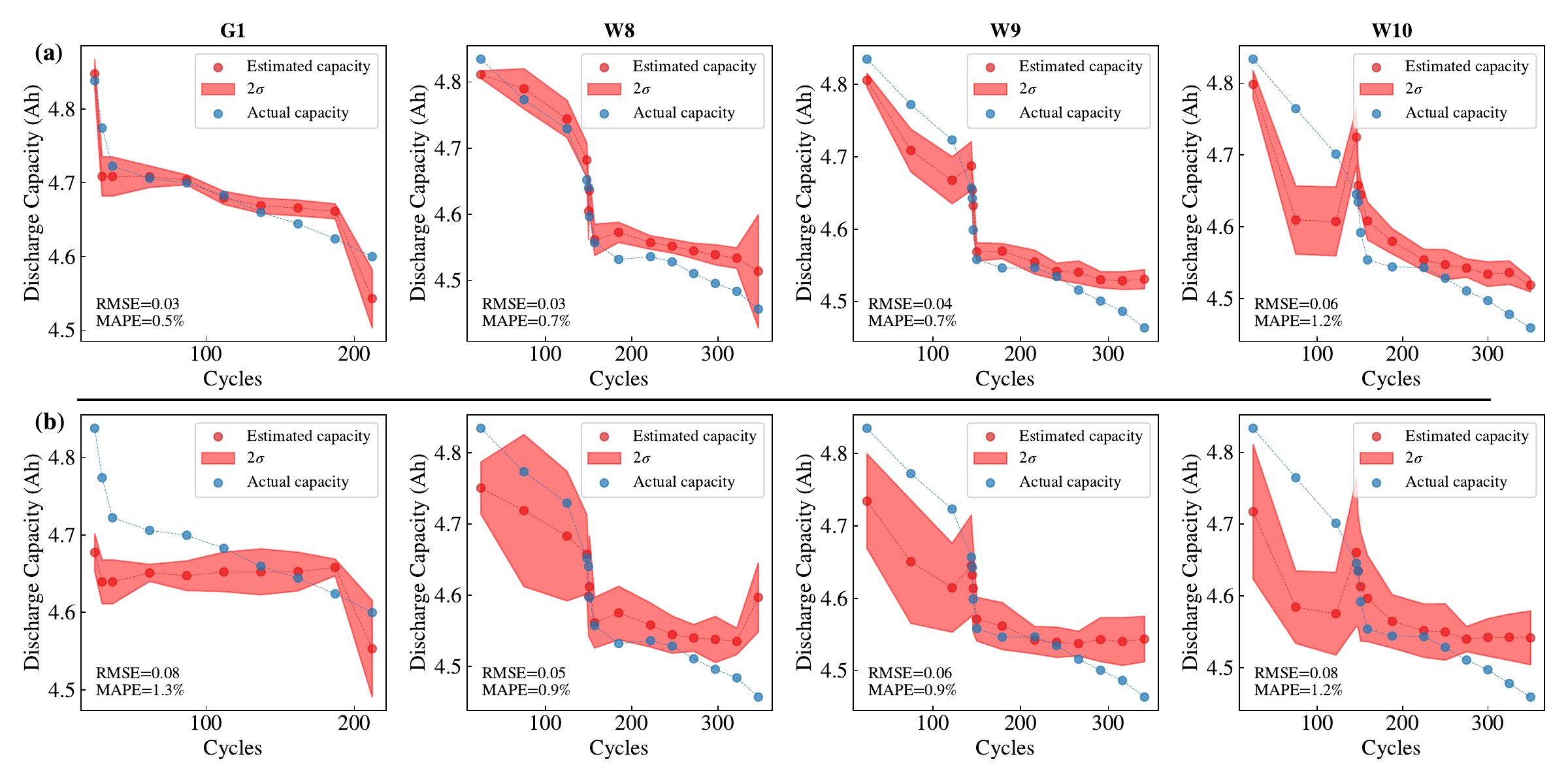}
  \caption{Results of the CNN capacity estimation model. (a) Model fine-tuned using pre-trained weights from the degradation SSL approach. (b) Model trained from scratch.}
  \label{fig:finetuned_result}
\end{figure}

The Stanford dataset contains only eight cells with approximately 100 data points measuring discharge capacity, making it challenging to train a capacity estimation model due to data limitations. Therefore, we leverage degradation-based self-supervised learning (SSL) as a pretext task to mitigate the sparsity of labeled information. The same CNN architecture and input features used in the degradation SSL model are employed to estimate actual discharge capacity. We compare the performance of a model fine-tuned with pre-trained weights against one trained from scratch.

In the capacity estimation test, 58 data points from six cells were used to train the downstream model, while the remaining 24 data points from two cells were reserved for testing. The model input consists of discharge voltage and current signals recorded within one hour of the corresponding cycle, with the discharge capacity of that cycle as the prediction target. Training was conducted using the Huber loss function, with a learning rate of 0.0001 and a weight decay coefficient of 0.001, until convergence.

Figure \ref{fig:finetuned_result} presents the prediction results for individual cells, and only W10 is from the testing set. To account for real-world variations in discharge data, multiple voltage-current segments sampled from different state-of-charge (SoC) intervals within the same cycle were used as model inputs, and the mean and standard deviation of the predicted values were computed. When using pre-trained weights, the model not only achieved higher prediction accuracy but also exhibited reduced estimation uncertainty caused by inconsistencies in signal sampling.

\section{Conclusion \& limitation}
This study presents a novel self-supervised learning framework for lithium-ion battery (LIB) health diagnostics, using the inherent irreversibility of battery degradation to design an effective pretext task. Unlike conventional supervised approaches that rely on labeled data from reference performance tests, the proposed self-supervised degradation learning method extracts meaningful health indicators directly from raw operating signals without requiring explicit capacity measurements. By learning the chronological order of the voltage and current profiles, the model successfully captures the battery degradation trends with a high correlation to the actual discharge capacity. Due to its independence from labeled training data, the proposed method offers a scalable and cost-effective solution for monitoring battery energy storage systems.

Several challenges remain to be addressed in future work. First, the Stanford datasets do not cover the full lifespan of lithium-ion batteries. For instance, the batteries in this dataset only degrade to approximately 90\% SoH on average. The ability to assess battery health near the end-of-life threshold, typically around 80\% SoH, requires further verification. Second, while the Stanford dataset use UDDS to replicate aging phenomena observed in electric vehicles, it is still conducted in a controlled laboratory environment. Acquiring real-world electric vehicle battery aging data would further optimize the evaluation process and enhance the commercial applicability of the proposed method.

\bibliographystyle{unsrt}  
\bibliography{main}  

\begin{thebibliography}{10}

\bibitem{sanguesa2021review}
Julio~A Sanguesa, Vicente Torres-Sanz, Piedad Garrido, Francisco~J Martinez, and Johann~M Marquez-Barja.
\newblock A review on electric vehicles: Technologies and challenges.
\newblock {\em Smart Cities}, 4(1):372--404, 2021.

\bibitem{liang2019review}
Yeru Liang, Chen-Zi Zhao, Hong Yuan, Yuan Chen, Weicai Zhang, Jia-Qi Huang, Dingshan Yu, Yingliang Liu, Maria-Magdalena Titirici, Yu-Lun Chueh, et~al.
\newblock A review of rechargeable batteries for portable electronic devices.
\newblock {\em InfoMat}, 1(1):6--32, 2019.

\bibitem{che2023health}
Yunhong Che, Xiaosong Hu, Xianke Lin, Jia Guo, and Remus Teodorescu.
\newblock Health prognostics for lithium-ion batteries: mechanisms, methods, and prospects.
\newblock {\em Energy \& Environmental Science}, 16(2):338--371, 2023.

\bibitem{birkl2017degradation}
Christoph~R Birkl, Matthew~R Roberts, Euan McTurk, Peter~G Bruce, and David~A Howey.
\newblock Degradation diagnostics for lithium ion cells.
\newblock {\em Journal of Power Sources}, 341:373--386, 2017.

\bibitem{chen2021review}
Yuqing Chen, Yuqiong Kang, Yun Zhao, Li~Wang, Jilei Liu, Yanxi Li, Zheng Liang, Xiangming He, Xing Li, Naser Tavajohi, et~al.
\newblock A review of lithium-ion battery safety concerns: The issues, strategies, and testing standards.
\newblock {\em Journal of Energy Chemistry}, 59:83--99, 2021.

\bibitem{an2016state}
Seong~Jin An, Jianlin Li, Claus Daniel, Debasish Mohanty, Shrikant Nagpure, and David~L Wood~III.
\newblock The state of understanding of the lithium-ion-battery graphite solid electrolyte interphase (sei) and its relationship to formation cycling.
\newblock {\em Carbon}, 105:52--76, 2016.

\bibitem{rahman2024exploring}
Tuhibur Rahman and Talal Alharbi.
\newblock Exploring lithium-ion battery degradation: A concise review of critical factors, impacts, data-driven degradation estimation techniques, and sustainable directions for energy storage systems.
\newblock {\em Batteries}, 10(7):220, 2024.

\bibitem{du2024side}
Hao Du, Yadong Wang, Yuqiong Kang, Yun Zhao, Yao Tian, Xianshu Wang, Yihong Tan, Zheng Liang, John Wozny, Tao Li, et~al.
\newblock Side reactions/changes in lithium-ion batteries: mechanisms and strategies for creating safer and better batteries.
\newblock {\em Advanced Materials}, 36(29):2401482, 2024.

\bibitem{severson2019data}
Kristen~A Severson, Peter~M Attia, Norman Jin, Nicholas Perkins, Benben Jiang, Zi~Yang, Michael~H Chen, Muratahan Aykol, Patrick~K Herring, Dimitrios Fraggedakis, et~al.
\newblock Data-driven prediction of battery cycle life before capacity degradation.
\newblock {\em Nature Energy}, 4(5):383--391, 2019.

\bibitem{chai2021deep}
Junyi Chai, Hao Zeng, Anming Li, and Eric~WT Ngai.
\newblock Deep learning in computer vision: A critical review of emerging techniques and application scenarios.
\newblock {\em Machine Learning with Applications}, 6:100134, 2021.

\bibitem{lim2021time}
Bryan Lim and Stefan Zohren.
\newblock Time-series forecasting with deep learning: a survey.
\newblock {\em Philosophical Transactions of the Royal Society A}, 379(2194):20200209, 2021.

\bibitem{fan2020novel}
Yaxiang Fan, Fei Xiao, Chaoran Li, Guorun Yang, and Xin Tang.
\newblock A novel deep learning framework for state of health estimation of lithium-ion battery.
\newblock {\em Journal of Energy Storage}, 32:101741, 2020.

\bibitem{hsu2022deep}
Chia-Wei Hsu, Rui Xiong, Nan-Yow Chen, Ju~Li, and Nien-Ti Tsou.
\newblock Deep neural network battery life and voltage prediction by using data of one cycle only.
\newblock {\em Applied Energy}, 306:118134, 2022.

\bibitem{lu2022battery}
Jiahuan Lu, Rui Xiong, Jinpeng Tian, Chenxu Wang, Chia-Wei Hsu, Nien-Ti Tsou, Fengchun Sun, and Ju~Li.
\newblock Battery degradation prediction against uncertain future conditions with recurrent neural network enabled deep learning.
\newblock {\em Energy Storage Materials}, 50:139--151, 2022.

\bibitem{dos2021lithium}
Gon{\c{c}}alo Dos~Reis, Calum Strange, Mohit Yadav, and Shawn Li.
\newblock Lithium-ion battery data and where to find it.
\newblock {\em Energy and AI}, 5:100081, 2021.

\bibitem{lombardo2021artificial}
Teo Lombardo, Marc Duquesnoy, Hassna El-Bouysidy, Fabian {\AA}r{\'e}n, Alfonso Gallo-Bueno, Peter~Bj{\o}rn J{\o}rgensen, Arghya Bhowmik, Arnaud Demorti{\`e}re, Elixabete Ayerbe, Francisco Alcaide, et~al.
\newblock Artificial intelligence applied to battery research: hype or reality?
\newblock {\em Chemical reviews}, 122(12):10899--10969, 2021.

\bibitem{rani2023self}
Veenu Rani, Syed~Tufael Nabi, Munish Kumar, Ajay Mittal, and Krishan Kumar.
\newblock Self-supervised learning: A succinct review.
\newblock {\em Archives of Computational Methods in Engineering}, 30(4):2761--2775, 2023.

\bibitem{hannan2021deep}
Mahammad~A Hannan, Dickson~NT How, MS~Hossain Lipu, M~Mansor, Pin~Jern Ker, ZY~Dong, KSM Sahari, Sieh~Kiong Tiong, Kashem~M Muttaqi, TM~Indra Mahlia, et~al.
\newblock Deep learning approach towards accurate state of charge estimation for lithium-ion batteries using self-supervised transformer model.
\newblock {\em Scientific reports}, 11(1):19541, 2021.

\bibitem{wang2024lithium}
Tianyu Wang, Zhongjing Ma, Suli Zou, Zhan Chen, and Peng Wang.
\newblock Lithium-ion battery state-of-health estimation: A self-supervised framework incorporating weak labels.
\newblock {\em Applied Energy}, 355:122332, 2024.

\bibitem{luo2024remaining}
Hong Luo, Jiang Yu, and Penghua Li.
\newblock Remaining useful life prediction of towards lithium-ion battery capacity region aggregated representation.
\newblock In {\em 2024 36th Chinese Control and Decision Conference (CCDC)}, pages 5238--5244. IEEE, 2024.

\bibitem{pozzato2022lithium}
Gabriele Pozzato, Anirudh Allam, and Simona Onori.
\newblock Lithium-ion battery aging dataset based on electric vehicle real-driving profiles.
\newblock {\em Data in brief}, 41:107995, 2022.

\bibitem{gilles2013empirical}
Jerome Gilles.
\newblock Empirical wavelet transform.
\newblock {\em IEEE transactions on signal processing}, 61(16):3999--4010, 2013.

\bibitem{al2023capacity}
Maher Al-Greer, Imran Bashir, et~al.
\newblock Capacity estimation of lithium-ion batteries based on adaptive empirical wavelet transform and long short-term memory neural network.
\newblock {\em Journal of Energy Storage}, 70:108046, 2023.

\bibitem{daubechies1992ten}
Ingrid Daubechies.
\newblock {\em Ten lectures on wavelets}.
\newblock SIAM, 1992.

\bibitem{liu2021dynamic}
Fang Liu, Rong Xu, Yecun Wu, David~Thomas Boyle, Ankun Yang, Jinwei Xu, Yangying Zhu, Yusheng Ye, Zhiao Yu, Zewen Zhang, et~al.
\newblock Dynamic spatial progression of isolated lithium during battery operations.
\newblock {\em Nature}, 600(7890):659--663, 2021.

\bibitem{hek2016deep}
M~HeK, Q~RenS, et~al.
\newblock Deep residual learning for image recognition.
\newblock In {\em 2016 IEEE Conference on Computer Vision and Pattern Recognition}, pages 770--778, 2016.

\bibitem{szegedy2016rethinking}
Christian Szegedy, Vincent Vanhoucke, Sergey Ioffe, Jon Shlens, and Zbigniew Wojna.
\newblock Rethinking the inception architecture for computer vision.
\newblock In {\em Proceedings of the IEEE conference on computer vision and pattern recognition}, pages 2818--2826, 2016.

\bibitem{loshchilov2017decoupled}
Ilya Loshchilov and Frank Hutter.
\newblock Decoupled weight decay regularization.
\newblock {\em arXiv preprint arXiv:1711.05101}, 2017.

\bibitem{reichert2013influence}
M~Reichert, D~Andre, A~R{\"o}smann, P~Janssen, H-G Bremes, DU~Sauer, S~Passerini, and M~Winter.
\newblock Influence of relaxation time on the lifetime of commercial lithium-ion cells.
\newblock {\em Journal of Power Sources}, 239:45--53, 2013.

\end{thebibliography}

\end{document}